\pdfoutput=1

\documentclass[11pt]{article}

\usepackage[final]{acl}

\usepackage{times}
\usepackage{latexsym}

\usepackage[T1]{fontenc}

\usepackage[utf8]{inputenc}

\usepackage{microtype}

\usepackage{inconsolata}

\usepackage{graphicx}

\usepackage{multirow}
\usepackage{booktabs}
\usepackage{array} 
\usepackage{pifont}
\usepackage{adjustbox}

\title{DrKGC: Dynamic Subgraph Retrieval-Augmented LLMs for Knowledge Graph Completion across General and Biomedical Domains}

\author{
 \textbf{Yongkang Xiao\textsuperscript{1}},
 \textbf{Sinian Zhang\textsuperscript{1}},
 \textbf{Yi Dai\textsuperscript{2}},
 \textbf{Huixue Zhou\textsuperscript{1}},
 \textbf{Jue Hou\textsuperscript{1}},
 \textbf{Jie Ding\textsuperscript{1}},
 \textbf{Rui Zhang\textsuperscript{1}\thanks{Corresponding author.}}
\\
\\
 \textsuperscript{1}University of Minnesota, Minneapolis, MN, USA \\
 \textsuperscript{2}University of Michigan, Ann Arbor, MI, USA 
\\
\texttt{\{xiao0290,ruizhang\}@umn.edu}
}

\begin{document}
\maketitle

\begin{abstract}

Knowledge graph completion (KGC) aims to predict missing triples in knowledge graphs (KGs) by leveraging existing triples and textual information. Recently, generative large language models (LLMs) have been increasingly employed for graph tasks. However, current approaches typically encode graph context in textual form, which fails to fully exploit the potential of LLMs for perceiving and reasoning about graph structures. To address this limitation, we propose \textbf{DrKGC} (\textbf{D}ynamic Subgraph \textbf{R}etrieval-Augmented LLMs for \textbf{K}nowledge \textbf{G}raph \textbf{C}ompletion). DrKGC employs a flexible lightweight model training strategy to learn structural embeddings and logical rules within the KG. It then leverages a novel bottom-up graph retrieval method to extract a subgraph for each query guided by the learned rules. Finally, a graph convolutional network (GCN) adapter uses the retrieved subgraph to enhance the structural embeddings, which are then integrated into the prompt for effective LLM fine-tuning. Experimental results on two general domain benchmark datasets and two biomedical datasets demonstrate the superior performance of DrKGC. Furthermore, a realistic case study in the biomedical domain highlights its interpretability and practical utility.\footnote{The code is available at \url{https://github.com/TheYKXiao/DrKGC}.}


\end{abstract}

\section{Introduction}

Knowledge graphs (KGs) are structured representations of real-world facts, typically formulated as a set of triples that consist of entities and their relationships \citep{nickel2015review,ji2021survey}. Biomedical Knowledge Graphs (BKGs) are specialized forms of KGs tailored to the biomedical domain. In a BKG, nodes represent biomedical entities—such as molecules, diseases, and genes—while edges capture various relationships among these entities, typically through functional predicates relevant to the biomedical domain (e.g., “treats,” “inhibits,” and “causes”) \citep{walsh2020biokg}. BKGs have proved instrumental in numerous biological tasks, including drug repurposing, side-effect prediction, and drug–drug interaction detection \citep{himmelstein2017systematic, zitnik2018modeling, lin2020kgnn}.

BKGs, like other KGs, often suffer from incompleteness, typically manifested as missing edges between nodes \citep{chen2020knowledge} . This incompleteness may arise because (1) the facts are absent from the data source, or (2) they remain undiscovered by humans. Such issues are particularly prevalent in BKGs, as their data primarily originates from experimental results, clinical trials, and scientific literature.

Inferring missing facts in knowledge graphs has led to the development of a wide range of Knowledge Graph Completion (KGC) models. These include structure-based methods (e.g., TransE~\citep{bordes2013translating} and R-GCN~\citep{schlichtkrull2018modeling}), rule-based methods (e.g., Neural-LP~\citep{yang2017differentiable}), and text-based methods (e.g., KG-BERT~\citep{yao2019kg}).
Recently, the advent of generative large language models (LLMs) has given rise to a new class of generation-based KGC approaches. Unlike traditional text-based methods that encode entity and relation descriptions into fixed embeddings, these approaches leverage LLMs to generate missing triples in a sequence-to-sequence manner, often relying on prompting or fine-tuning strategies (e.g., KICGPT \citep{wei2024kicgpt}, KoPA \citep{zhang2024making}). Although generation-based methods have shown promise in KGC, they face several key limitations: \ding{182} \textbf{Structural Information Loss}: These methods often fail to preserve the rich structural information inherent in knowledge graphs. While graph paths or subgraphs can be encoded as text prompts, overly long inputs introduce noise and increase computational costs. \ding{183} \textbf{Static Embedding Limitations}: Incorporating structural embeddings into LLMs offers a partial solution but remains limited, as such embeddings are static and do not adapt to the query-specific context or dynamic subgraph structure. \ding{184} \textbf{Generic Responses from LLMs}: In the absence of additional constraints, LLMs tend to generate generic predictions influenced by pretraining data. This is especially problematic in biomedical KGs, where high-degree entities and many-to-many relations make multiple answers plausible—yet not all are contextually correct or desirable.
 
To tackle these challenges, we propose \textbf{D}ynamic Subgraph \textbf{R}etrieval-Augmented LLMs for \textbf{K}nowledge \textbf{G}raph \textbf{C}ompletion (\textbf{DrKGC}). Our approach begins by converting incomplete triples into natural language questions using an automatically generated template lexicon. It then employs lightweight models to learn global structural embeddings of entities, discover logical rules and coarsely rank candidate entities based on their relevance to the query. This anchors the reasoning process in semantic and structural context without requiring long or noisy text prompts (\ding{182}).
To overcome the limitations of static embeddings (\ding{183}), DrKGC dynamically constructs a query-specific subgraph using retrieved candidates and learned logical rules. This enables the model to focus on relevant local structures and incorporate adaptive, context-aware structural cues during inference.
Finally, to mitigate the risk of generic or irrelevant responses (\ding{184}), DrKGC restricts the output space by explicitly defining a candidate entity set. The prompt is enriched with both global and local graph signals, guiding the LLM to generate contextually grounded and targeted predictions, especially in cases involving complex, many-to-many relations.
 

The key contributions of our work are as follows:
\begin{itemize}
    \item We propose DrKGC, a novel and flexible framework for knowledge graph completion that effectively supports both general KGs and domain BKGs.
    \item We develop two critical components of DrKGC to effectively integrate graph-structural information into the generative model. Specifically, we extend the standard retrieval-augmented generation to the graph scenario where we leverage logical rules to obtain a local subgraph that represents entities of potential interest. Then, we develop a technique that applies graph convolutional networks to the retrieved subgraphs to further generate local embeddings of entities, effectively supplying structural information for LLM-based prediction. 
    \item We perform comprehensive experiments on both benchmark datasets and biomedical use cases to evaluate the performance of DrKGC and show its significant improvement over state-of-the-art baseline approaches. We further conduct a biomedical case study on drug repurposing to demonstrate the practical applicability of DrKGC.
\end{itemize}

\section{Related Work}
\subsection{Structure-based Methods}


Structure-based KGC methods leverage the structural information of nodes and edges in large heterogeneous graphs. Early methods learn low-dimensional embeddings for entities and relations based on individual triples—for example, TransE \citep{bordes2013translating} views a relation as a translation from the subject to the object, while RotatE \citep{sun2019rotate} extends TransE into a complex space to model symmetric relations. Semantic matching approaches (e.g., ComplEx \citep{trouillon2016complex}, DistMult \citep{yang2014embedding}) compute the similarity of entity and relation representations. However, these triple-based methods handle each triple independently and ignore higher-order neighborhood information. To address this, GNN-based methods, such as R-GCN \citep{schlichtkrull2018modeling} and CompGCN \citep{vashishth2019composition-based}, introduce message passing and neighborhood aggregation to incorporate multi-hop context.

\subsection{Rule-based Methods}


Because two entities in a KG may be linked by a few one-hop paths but numerous multi-hop paths, rule-based methods have emerged to learn probabilistic logic rules from these relation paths for inferring missing triples. For example, Neural-LP \citep{yang2017differentiable} offers an end-to-end differentiable framework that jointly learns the parameters and structures of first-order logical rules by combining a neural controller with attention and memory, composing differentiable TensorLog operations. NCRL \citep{cheng2023neuralcompositionalrulelearning} learns logical rules by splitting rule bodies into smaller parts, encoding them via a sliding window, and then merging them recursively with an attention mechanism, achieving efficient and scalable reasoning.

\subsection{Text-based Methods}


Knowledge graphs often include extensive textual information, such as names and descriptions of entities and relations, which text-based methods can exploit using pre-trained language models (PLMs) to predict missing triples. For example, KG-BERT \citep{yao2019kg} computes triple scores by feeding the text of head entities, relations, and tail entities into a BERT model. SimKGC \citep{wang2022simkgc} applies contrastive learning with three types of negative samples to build more discriminative KGC models. KGLM \citep{youn2022kglm} augments PLMs with entity and relation embeddings that capture KG structural information, further improving link prediction performance.

\subsection{Generation-based Methods}

With the rise of generative large language models (LLMs), generation-based approaches have gained attention by reformulating KGC into a sequence-to-sequence text generation task. These methods still rely on textual information from KGs, but they reframe a KGC query as a natural language question, prompt the LLM for an answer, and map that output back to KG entities. For example, KICGPT \citep{wei2024kicgpt} introduces an in-context learning strategy that uses explicit instructions to guide LLM reasoning.  KG-LLM \citep{yao2025exploring} applies LLMs for triple classification and relation prediction, highlighting their adaptability to diverse KGC subtasks. KoPA \citep{zhang2024making} introduces the Knowledge Prefix Adapter to integrate pre-trained structural embeddings into LLMs to enhance structure-aware reasoning. From a prompting perspective, LPNL \citep{bi2024lpnl} uses a two-stage sampling and divide-and-conquer method for scalable link prediction via prompts. KC-GenRea \citep{wang2024kc} reformulates KGC as a LLM-based re-ranking task, and DIFT \citep{liu2024finetuninggenerativelargelanguage} implements KGC using discriminant instructions to finetune LLMs.

\subsection{Biomedical Knowledge Graph Completion}


BKGs have gained substantial attention for modeling structured knowledge in complex biomedical systems. Representative BKGs include Hetionet \citep{himmelstein2017systematic}, unifying 29 databases into a single network, PharmKG \citep{zheng2021pharmkg}, integrating 6 databases plus text-mined knowledge, and PrimeKG \citep{chandak2023building}, a precision medicine–focused graph consolidating 20 resources. For BKGs, KGC is essential in identifying missing triples to generate new hypotheses—for example, ICInet \citep{zhao2023biological} integrates GNNs, biological KGs, and gene expression profiles to predict cancer immunotherapy outcomes, while FuseLinker \citep{xiao2024fuselinker} fuses pre-trained LLM text embeddings with Poincaré graph embeddings for improved GNN-based link prediction in drug repurposing.

\section{Methodology}
In this section, we introduce the proposed DrKGC. We begin with the preliminary and an overview, followed by a detailed description of each component.

\subsection{Preliminary}
\noindent \textbf{Knowledge Graph (KG).} 
A KG (or BKG) can be represented as a directed multigraph, $\mathcal{G}=(\mathcal{E},\mathcal{R},\mathcal{T})$, where $\mathcal{E}$ is the set of entities, $\mathcal{R}$ is the set of relations and $\mathcal{T}=\{(h,r,t)|h,t \in \mathcal{E}, r \in \mathcal{R}\}$ is the set of triples. Each triple $(h,r,t)$, with $h$ and $t$ denoting the head and tail entities, and $r$ representing the relation between them, describes a fact in the KG.

\noindent \textbf{Knowledge Graph Completion (KGC).} 
KGC aims to infer novel or missing triples from those already present in the graph. Let triples $\{(h',r',t')|h',t' \in \mathcal{E}, r' \in \mathcal{R}\}$, with $(h',r',t') \notin \mathcal{T}$, represent facts that are unobserved in the KG.  In this work, we cast KGC as the task of identifying missing entities in incomplete triples $(?,r_q,t_q)$ and $(h_q,r_q,?)$, which are referred to as head prediction and tail prediction, respectively. Here, $h_q$ or $t_q$ is the query entity, and $r_q$ is the query relation.

\subsection{Overview}

\begin{figure*}[htbp]
    \centering
    \includegraphics[width=1\linewidth]{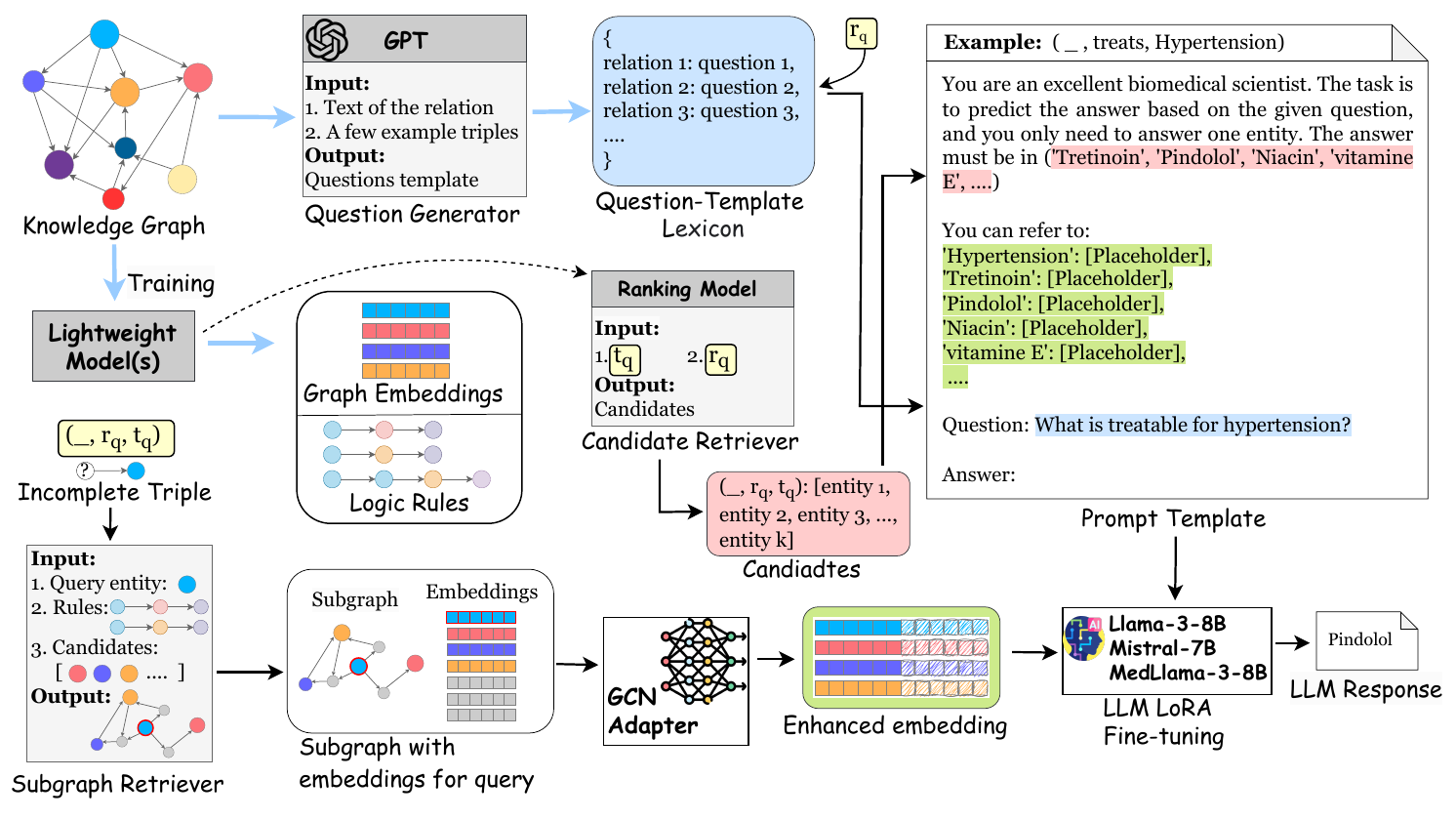} %
    \caption{Overview of the DrKGC framework. Light-blue arrows denote the dataset-level workflow (run once per KG); black arrows denote the per-triple workflow (run for each incomplete triple).} 
    \label{fig:framework} 
\end{figure*}


For simplicity, we only consider the head prediction scenario for illustration. Figure~\ref{fig:framework} illustrates the overall framework of DrKGC. DrKGC first employs a Question Generator to convert the incomplete triples $(?,r_q,t_q)$ into well-formed question $Q$. Then, a pre-trained lightweight model scores each entity $\{e \in \mathcal{E} \mid (e, r_q, t_q) \notin \mathcal{T}\}$ for $(?,r_q,t_q)$, and selects the top $k$ entities, where $k$ is a hyperparameter, to form a candidates set $C=[e_1,e_2,e_3,\dots,e_k]$. Subsequently, Subgraph Retriever retrieves a subgraph $G$ based on the query entity $t_q$, all entities in $C$ and the logic rules of $r_q$. A GCN-based adapter then leverages $G$ to refine the embeddings of the $t_q$ and the entities in $C$. Finally, the LLM selects the most plausible entity from the $C$, using both its own knowledge and the structured embeddings, to answer the question $Q$.

\subsection{Question Generator}
To more accurately express the relations in the KG and convey the specific functional semantics of relations in BKGs, we reformulate the KGC task into a question-answering paradigm that aligns with LLMs. To achieve this, we introduce a simple yet effective approach comprising two main stages: Template Generation and Question Generation.

\noindent \textbf{Template Generation.} 
For each KG, we conduct a one-time template generation process using GPT's few-shot in-context learning. Specifically, GPT-o1 is provided with a relation's name, its textual description, and a small set of sample triples, and is then instructed to generate a corresponding question template (with a placeholder for the query entity) via pattern induction. After processing each relation, we compile a question-template lexicon $L$ (distinguishing between head and tail predictions) covering the entire relations set.  Appendix~\ref{app:template_lexicon} shows the lexicon for WN18RR as an example.

\noindent \textbf{Question Generation.}
After obtaining the lexicon $L$, we map the query relation $r_q$ to its corresponding question template and then place the query entity $t_q$ into the placeholder position to generate the complete question $Q$, which can be expressed as $Q=P(L(r_q),t_q)$. 

\subsection{Candidate Retriever}
To mitigate the issues of an excessively large search space, limited LLM input capacity, and the tendency of LLMs to produce generic responses, we constrain the LLM's input and output using candidates sets. Following prior works (e.g., \cite{zhang2024making,wei2024kicgpt,liu2024finetuninggenerativelargelanguage}), we employ lightweight KGC methods to produce entity rankings, which are then used to collect candidate entities.

\noindent \textbf{Lightweight Model Training.}
Unlike previous work, we require more than just entity rankings. Therefore, we train lightweight structure-based models to obtain the structural embeddings of entities and lightweight rule-based models to learn the logical rules of relations in the KG, which guide the subsequent subgraph retrieval. By “lightweight,” we refer to simpler, more resource-efficient approaches that do not rely on large-scale pretraining. This process is inherently flexible and any advanced method that can generate structured embeddings and perform rule mining may serve as a substitute. In our implementation, we focus on leveraging open-source methods that have demonstrated strong performance on the KGC task. The best structure-based model, $M_S$, generates embeddings for all entities, which we denote as the global embeddings $\mathbf{E}_{global} = \{ \mathbf{e}_{global} \mid e \in \mathcal{E} \}$. The best rule-based model, $M_R$ extracts logic rules for each relation. For every relation $r \in \mathcal{R}$, we denote the corresponding set of rules as $\mathcal{L}_r$.

\noindent \textbf{Candidates Collection.}
The best performing lightweight model is used to coarsely rank the candidates. Specifically, for head prediction with $r_q$ and $t_q$, we replace the head with each entity $\{e \in \mathcal{E} \mid (e, r_q, t_q) \notin \mathcal{T}\}$ for incomplete triple $(?,r_q,t_q)$, and compute a likelihood score for each resulting triple, which reflects its plausibility. The replaced entities in top $k$ triples are then selected to form the candidates set $C=[e_1,e_2,e_3,\dots,e_k]$.

\subsection{Dynamic Subgraph RAG}
Retrieval-augmented generation (RAG) integrates retrieval-based methods with generative models to enhance the quality and accuracy of generated text \citep{lewis2020retrieval}. Inspired by this idea, we propose a dynamic subgraph RAG strategy tailored for KGC tasks, which comprises two key components: Dynamic Subgraph Retrieval and Structure-Aware Embedding Enhancement. 

\noindent \textbf{Dynamic Subgraph Retrieval.} 
To enable the LLM to select the correct answer from $C$ based on the query entity  $t_q$ and query relation $r_q$, it is crucial to retrieve an informative subgraph to augment the graph context. To this end, we propose a bottom-up dynamic subgraph retrieval scheme, which is dynamic in that it does not mechanically retrieve the subgraph solely based on the $t_q$ and $r_q$, but rather adapts to variable candidates sets. Specifically, we first retrieve the shortest paths connecting each $e \in C$ to the $t_q$ to  ensure that both the $t_q$ and all candidate entities $e \in C$ are included in the subgraph $G$ and guarantee connectivity. Next, we sort the logical rules in $\mathcal{L}_{r_q}$ by their assigned confidence 
scores and sequentially use them to search the paths from the $e \in C$ to $t_q$, thereby enriching the subgraph. This process continues until the number of triples reaches a preset threshold $\tau$, which serves to constrain the subgraph's size. Finally, if the number of triples remains below $\tau$ after these steps, we augment the subgraph with additional triples connected to other entities from $e \in C$ and $t_q$ via the $r_q$ and its logical rules. More details are provided in Appendix~\ref{app:subgraph retrieval strategy}.

\noindent \textbf{Structure-Aware Embedding Enhancement.} 

Unlike traditional RAG, integrating structured subgraphs directly into the prompts is challenging. Even if described in text, much of the structural information is lost, and the text may be excessively long due to the richness of the subgraphs. To overcome this limitation, we exploit the subgraph’s structural information to vectorize the graph context. We refer to the resulting embeddings as local embeddings $\mathbf{E}_{local} = \{ \mathbf{e}_{local} \mid e \in \mathcal{E} \}$.

To obtain local embeddings and enhance the overall structural representation, we design a graph convolutional network (GCN)-based adapter. It comprises a low-dimensional relational GCN and a subsequent adapter that projects the resulting vectors to the LLM input layer’s dimensionality. Specifically, for each query subgraph, the GCN is initialized with the global embeddings of all entities and then updates these representations via the neighborhood aggregation mechanism to produce the local embeddings. We concatenate the global and local embeddings to form the final enhanced structural embedding, i.e., $\mathbf{e}_{enhance} = [\mathbf{e}_{global};\mathbf{e}_{local}]$. To reduce computational overhead for graphs, GCN computations are performed in a low-dimensional space. Consequently, we employ an adapter to map the resulting structural embeddings to the LLM input dimension for seamless integration. During LoRA fine-tuning, we allow gradients to flow through the entire model, including the GCN adapter.

\subsection{Prompt Template}
Appendix~\ref{app:prompt_template} presents the detailed prompt template. In summary,  for each queried incomplete triple, our prompt comprises the following components: the instruction $I$ for KGC; the candidates set $C$; special \texttt{[Placeholder]} tokens for the structured embeddings, which are replaced by the actual enhanced structural embeddings of $t_q$ and each $e \in C$ after token vectorization; and the question $Q$ generated by the Question Generator.

\section{Experiments}
\subsection{Experiment Setup}
\textbf{Dataset.} 
We evaluate our proposed method on two benchmark KG datasets, WN18RR \citep{dettmers2018convolutional} and FB15k-237 \citep{toutanova2015representing}, and two widely used BKG datasets, PharmKG \citep{zheng2021pharmkg} and PrimeKG \citep{chandak2023building}. Dataset statistics, detailed descriptions and processing procedures are provided in Appendix~\ref{app:dataset}.

\begin{table*}[!t]
\centering
\scriptsize                      
\setlength{\tabcolsep}{3pt}       
\renewcommand{\arraystretch}{0.9} 
\resizebox{0.89\textwidth}{!}{%
\begin{tabular}{l|lcccccccc}
\toprule
\multicolumn{2}{c}{\multirow{2}{*}{\centering Methods}} & \multicolumn{4}{c}{\textbf{WN18RR}} & \multicolumn{4}{c}{\textbf{FB15k-237}} \\ \cmidrule(lr){3-6} \cmidrule(lr){7-10}
\multicolumn{2}{c}{}                               & MRR   & Hits@1 & Hits@3 & Hits@10 & MRR   & Hits@1 & Hits@3 & Hits@10 \\ \midrule
\multirow{4}{*}{Structure-based} 
                          & TransE                    & 0.243 & 0.043  & 0.441  & 0.532  & 0.279 & 0.198 & 0.376 & 0.441  \\ 
                          & DistMult                  & 0.444 & 0.412  & 0.470  & 0.504  & 0.281 & 0.199 & 0.301 & 0.446  \\
                          & RotatE                    & 0.476 & 0.428  & 0.492  & 0.571  & 0.338 & 0.241 & 0.375 & 0.533  \\  
                          & CompGCN                   & 0.479 & 0.443  & 0.494  & 0.546  & 0.355 & 0.264 & 0.390 & 0.535  \\ \cmidrule(lr){1-10}
\multirow{3}{*}{Rule-based} 
                          & Neural-LP                 & 0.381 & 0.368  & 0.386  & 0.408  & 0.237 & 0.173 & 0.259 & 0.361  \\ 
                          & RLogic                    & 0.470 & 0.443  & --     & 0.537  & 0.310 & 0.203 & --    & 0.501  \\ 
                          & NCRL                      & 0.670 & 0.563  & --     & \textbf{0.850} & 0.300 & 0.209 & --    & 0.473  \\ \cmidrule(lr){1-10}
\multirow{3}{*}{Text-based} 
                          & KG-BERT                   & 0.216 & 0.041  & 0.302  & 0.524  & --    & --    & --    & 0.420  \\ 
                          & SimKGC                    & 0.671 & 0.595  & 0.719  & 0.802  & 0.336 & 0.249 & 0.362 & 0.511  \\ 
                          & KGLM                    & 0.467 & 0.330  & 0.538  & 0.741  & 0.298 & 0.200 & 0.314 & 0.468  \\                           
                          & GHN                       & 0.678 & 0.596  & 0.719  & \underline{0.821} & 0.339 & 0.251 & 0.364 & 0.518  \\  \cmidrule(lr){1-10}
\multirow{3}{*}{Generation-based} 
                          & KICGPT                    & 0.564 & 0.478  & 0.612  & 0.677  & 0.412 & 0.327 & 0.448 & 0.554  \\ 
                          & COSIGN                    & 0.641 & 0.610  & 0.654  & 0.715  & 0.368 & 0.315 & 0.434 & 0.520  \\  
                          & DIFT                      & \underline{0.686} & \underline{0.616} & \underline{0.730} & 0.806  & \underline{0.439} & \underline{0.364} & \underline{0.468} & \underline{0.586}  \\ \cmidrule(lr){1-10}
\multirow{2}{*}{Hybrid} 
                          & StAR                      & 0.551 & 0.459  & 0.594  & 0.732  & 0.365 & 0.266 & 0.404 & 0.562  \\
                          & CoLE                      & 0.587 & 0.532  & 0.608  & 0.694  & 0.389 & 0.294 & 0.430 & 0.574  \\ \cmidrule(lr){1-10}
\multicolumn{2}{c}{DrKGC (Ours)}              & \textbf{0.716} & \textbf{0.654} & \textbf{0.757} & 0.813  & \textbf{0.472} & \textbf{0.406} & \textbf{0.498} & \textbf{0.599}  \\ 
\midrule
\multicolumn{2}{c}{\multirow{2}{*}{\centering Methods}} & \multicolumn{4}{c}{\textbf{PharmKG}} & \multicolumn{4}{c}{\textbf{PrimeKG}} \\ \cmidrule(lr){3-6} \cmidrule(lr){7-10}
\multicolumn{2}{c}{}                               & MRR   & Hits@1 & Hits@3 & Hits@10 & MRR   & Hits@1 & Hits@3 & Hits@10 \\ \midrule
\multirow{5}{*}{Structure-based} 
                          & TransE                    & 0.091  & 0.034  & 0.092  & 0.198  & 0.281  & 0.194  & 0.315  & 0.451  \\ 
                          & RotatE                    & -      & -      & -      & -      & 0.382  & 0.285  & 0.419  & 0.588  \\ 
                          & DistMult                  & 0.063  & 0.024  & 0.058  & 0.133  & 0.212  & 0.148  & 0.238  & 0.341  \\ 
                          & ComplEx                   & 0.075  & 0.030  & 0.071  & 0.155  & 0.204  & 0.141  & 0.266  & 0.340  \\ 
                          & R-GCN                    & 0.067  & 0.027  & 0.062  & 0.139  & \underline{0.640}  & \underline{0.569}  & \underline{0.680}  & \underline{0.761}  \\ 
                          & HRGAT                    & \underline{0.154}  & \underline{0.075}  & \underline{0.172}  & \underline{0.315}  & 0.443  & 0.347  & 0.489  & 0.637 \\ 
\midrule
\multicolumn{2}{c}{DrKGC (Ours)}               & \textbf{0.266}  & \textbf{0.183}  & \textbf{0.293}  & \textbf{0.436}  & \textbf{0.658} & \textbf{0.592} & \textbf{0.705} & \textbf{0.770} \\ 
\bottomrule
\end{tabular}
}
\caption{Comparison of DrKGC (using Llama-3-8B) and baselines on WN18RR, FB15k-237, PharmKG and PrimeKG. For each metric, the best performance is highlighted in \textbf{bold}, and the second-best is \underline{underlined}.}
\label{tab:main_results}
\end{table*}

\noindent \textbf{Baseline Methods.}
For the KG and BKG datasets, we selected two sets of baselines.

(1) For the WN18RR and FB15k-237, we consider baselines spanning multiple categories: structure-based methods: TransE \citep{bordes2013translating}, DistMult \citep{yang2014embedding}, RotatE \citep{sun2019rotate} and CompGCN \citep{vashishth2019composition-based}; rule-based methods: Neural-LP \citep{yang2017differentiable}, RLogic \citep{cheng2022rlogic}, and NCRL \citep{cheng2023neuralcompositionalrulelearning}; text-based methods: KG-BERT \citep{yao2019kg}, SimKGC \citep{wang2022simkgc}, KGLM \citep{youn2022kglm} and GHN \citep{qiao-etal-2023-improving}; generation-based methods: KICGPT \citep{wei2024kicgpt}, COSIGN \citep{li2024cosign} and DIFT \citep{liu2024finetuninggenerativelargelanguage}; and hybrid methods: StAR \citep{wang2021structure} and CoLE \citep{10.1145/3511808.3557355}. The baseline comparisons in this paper are based on the reported performance values of these methods.

(2) Since textual information essential for text- and generation-based methods is often sparse and unbalanced in BKGs, we focus on structure-based methods on PharmKG and PrimeKG, including TransE, RotatE, DistMult, ComplEx, R-GCN \citep{schlichtkrull2018modeling}, and HRGAT \citep{10.1145/3545573}, all of which are widely used as baselines for link prediction tasks on BKGs. The baseline performance for PharmKG is taken from the values reported in the original PharmKG \citep{zheng2021pharmkg} paper; while for PrimeKG, the baseline comparisons were conducted by ourselves.

\noindent \textbf{Implementation Details.}
In the lightweight model training stage, we trained NCRL to mine logical rules for the four datasets. For global structural embeddings, we employed RotatE for WN18RR and FB15k-237, and HRGAT for PharmKG, with hyperparameters consistent with the original publications. For PrimeKG, we used R-GCN with our optimal hyperparameter settings to obtain global embeddings. For WN18RR and FB15k-237, we additionally utilize the ranking results from SimKGC and CoLE, whereas, for PharmKG and PrimeKG, we directly employ HRGAT and R-GCN for ranking. The candidates set size is set at 20. For the fine-tuning stage, we compared Llama-3-8B \citep{dubey2024llama}, Llama-3.2-3B \citep{dubey2024llama}, MedLlama-3-8B \citep{medllama3b} and Mistral-7B \citep{jiang2023mistral} as our LLMs. We employed LoRA for efficient parameter tuning, with the primary hyperparameters set to $r=32$, $\alpha=32$, a dropout rate of $0.1$ and a learning rate of $2 \times 10^{-4}$. Model performance was evaluated using ranking-based metrics, including Mean Reciprocal Rank (MRR) and Hits@1, Hits@3, and Hits@10 under the “filtered” setting \citep{bordes2013translating}.  Additional training details are in Appendix~\ref{app:training}.

All experiments were conducted on an AMD EPYC 7763 64-Core CPU, an NVIDIA A100-SXM4-40GB GPU (CUDA 12.4), and Rocky Linux 8.10.

\subsection{Main Results}


Table~\ref{tab:main_results} shows that DrKGC achieves state-of-the-art performance on WN18RR, FB15k-237, PharmKG, and PrimeKG across most metrics. On WN18RR, although DrKGC trails NCRL and GHN in Hits@10, it outperforms all generation-based methods. For example, compared with DIFT \citep{liu2024finetuninggenerativelargelanguage} which also uses SimKGC as a ranker, DrKGC improves MRR by $4.37\%$ and Hits@1 by $6.17\%$. The gap with the top-performing text-based GHN in Hits@10 is also minimal ($-0.97\%$).  

For FB15k-237, DrKGC outperforms all baselines on every metric, with improvements of $7.5\%$ in MRR and $11.4\%$ in Hits@1. Given FB15k-237’s diverse set of relations and semantic patterns, these results underscore the ability of DrKGC to capture heterogeneous relational structures and semantic nuances.   

On PharmKG and PrimeKG, DrKGC consistently outperforms all baselines, demonstrating that even though BKGs lack extensive text information and LLMs are not pre-trained on specialized biomedical corpora, DrKGC can still achieve strong results by leveraging semantic understanding together with structural embeddings.

\subsection{Ablation Studies}
We conducted ablation studies on all four datasets to assess the contribution of each component in DrKGC, with the results presented in Table~\ref{tab:combined_ablation}. In the first ablation study, we removed the rule restrictions during subgraph retrieval. The results show that DrKGC’s performance declined across all four datasets, with a more pronounced drop in KGs than in BKGs. In the second study, we eliminated local embeddings and relied solely on global embeddings as the structural reference for entities. This change also led to performance degradation on all datasets. In the third study, we removed the structural embeddings entirely, forcing the LLM to select the correct answer directly from the candidates set without any structural reference. The significant performance decline observed for both KGs and BKGs confirms the importance of incorporating structural information into LLM predictions. Finally, we omitted the question template and instead directly instructed the LLM to complete the incomplete triple. While it resulted in only a slight performance drop on KGs, it had a substantial impact on BKGs. This can be attributed to the fact that relations in BKGs are inherently functional and mechanistic; for instance, asking the LLM "What gene causes Parkinson's disease?" provides clearer instruction than simply prompting it to complete an incomplete triple such as (?, causes, Parkinson's disease).

\begin{table}[ht]
\centering
\resizebox{\columnwidth}{!}{%
\begin{tabular}{l|cc|cc}
\toprule
\multirow{2}{*}{w/o} & \multicolumn{2}{c|}{\textbf{WN18RR}} & \multicolumn{2}{c}{\textbf{FB15k-237}} \\ 
\cmidrule(lr){2-3} \cmidrule(lr){4-5}
                   & MRR (\(\Delta\%\)) & Hits@1 (\(\Delta\%\)) & MRR (\(\Delta\%\)) & Hits@1 (\(\Delta\%\)) \\ 
\midrule
rules             & 0.684 ( -4.47)   & 0.612 ( -6.42)    & 0.448 ( -5.08)   & 0.375 ( -7.64) \\
local embedding   & 0.676 ( -5.59)   & 0.596 ( -8.87)    & 0.439 ( -6.99)   & 0.361 ( -11.1) \\
embedding         & 0.669 ( -6.56)   & 0.582 ( -11.0)    & 0.433 ( -8.26)   & 0.351 ( -13.5) \\
question template & 0.711 ( -0.70)   & 0.647 ( -1.07)    & 0.469 ( -0.64)   & 0.401 ( -1.23) \\
\midrule
DrKGC (raw)       & 0.716           & 0.654            & 0.472           & 0.406 \\
\midrule
\multirow{2}{*}{w/o} & \multicolumn{2}{c|}{\textbf{PharmKG}} & \multicolumn{2}{c}{\textbf{PrimeKG}} \\ 
\cmidrule(lr){2-3} \cmidrule(lr){4-5}
                   & MRR (\(\Delta\%\)) & Hits@1 (\(\Delta\%\)) & MRR (\(\Delta\%\)) & Hits@1 (\(\Delta\%\)) \\ 
\midrule
rules             & 0.264 ( -0.75)   & 0.181 ( -1.09)    & 0.648 ( -1.52)   & 0.578 ( -2.36) \\
local embedding   & 0.261 ( -0.88)   & 0.176 ( -3.83)    & 0.631 ( -4.10)   & 0.539 ( -8.95) \\
embedding         & 0.260 ( -2.26)   & 0.174 ( -4.92)    & 0.619 ( -5.93)   & 0.510 ( -13.9) \\
question template & 0.258 ( -3.01)   & 0.172 ( -6.01)    & 0.613 ( -6.83)   & 0.510 ( -13.9) \\
\midrule
DrKGC (raw)       & 0.266           & 0.183            & 0.658           & 0.592 \\
\bottomrule
\end{tabular}%
}
\caption{Ablation study results on four datasets.}
\label{tab:combined_ablation}
\end{table}




\subsection{DrKGC under Complex Conditions}
\begin{figure}[htbp]
    \centering
    \includegraphics[width=1\linewidth]{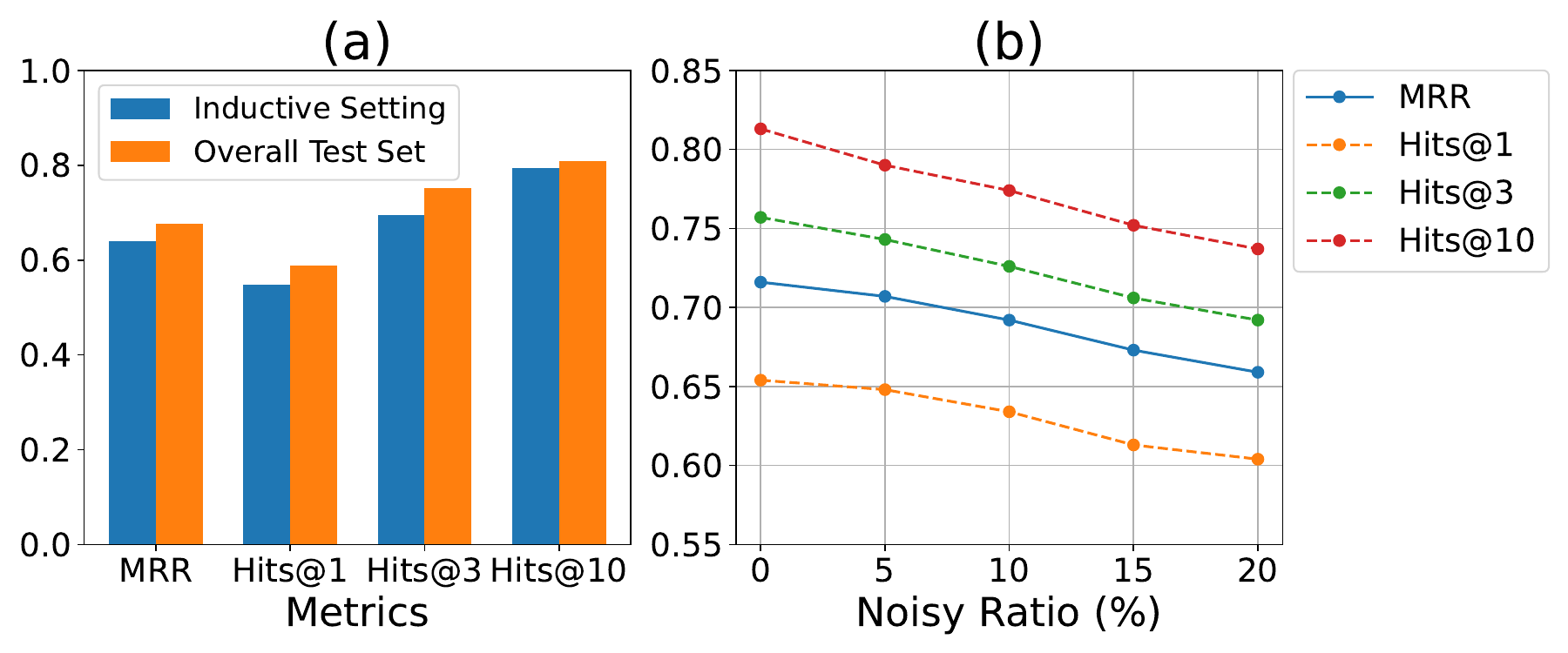} %
    \caption{Robustness evaluation on WN18RR. (a) Comparison of evaluation metrics under the inductive setting versus the overall test set. (b) Impact of proportional noise addition on model performance.} 
    \label{fig:robust} 
\end{figure}

To further verify DrKGC’s robustness, we evaluated both its inductive prediction capability and its resilience under noisy conditions on WN18RR. Specifically, for the inductive setting, we extracted all test triples whose entities or relation never appear in the training set and measured DrKGC’s performance on those unseen-entity cases.  For the noise experiment, we replaced a fixed proportion of triples in the training set with random negative triples and then assessed the resulting impact on DrKGC’s metrics. The results are summarized in Figure~\ref{fig:robust}. Under the inductive setting, our model experiences only modest performance drops (MRR: $-5.4 \%$; Hits@1: $-6.7 \%$), and even when injecting noise into $20 \%$ of the KG, the reductions in MRR and Hits@1 remain limited to $-7.9 \%$ and $-7.6 \%$, respectively, demonstrating DrKGC’s robustness.

\subsection{Subgraph Size Sensitivity Analysis}
\begin{figure}[htbp]
    \centering
    \includegraphics[width=0.95\linewidth]{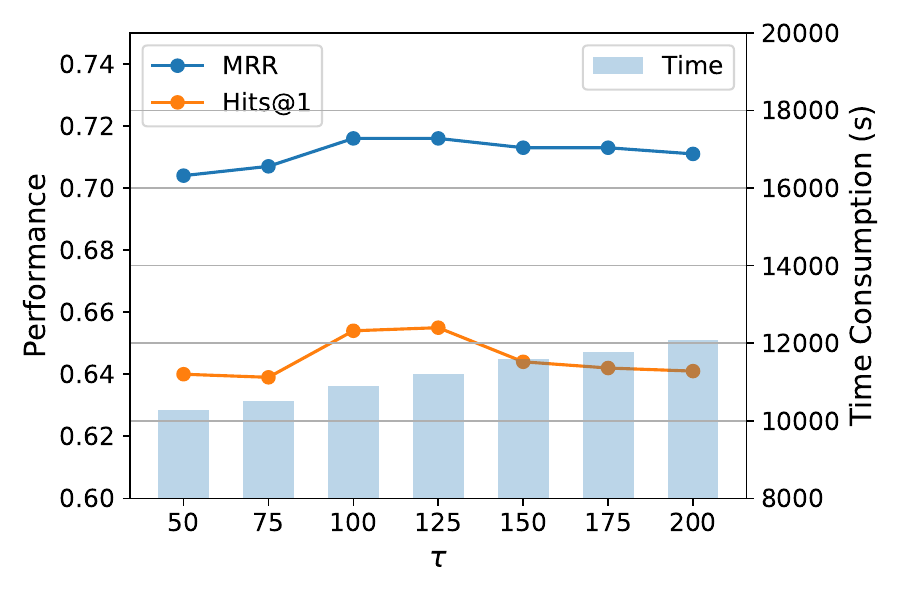} %
    \caption{Impact of $\tau$ on DrKGC Performance and time consumption on WN18RR.} 
    \label{fig:tau} 
\end{figure}

We examine how varying the threshold $\tau$, which controls subgraph size, affects model performance and efficiency on WN18RR. The  results are presented in Figure~\ref{fig:tau}. As $\tau$ increases, model performance initially improves and then declines, with optimal results observed at $\tau=100$ (our chosen hyperparameter) or $125$; conversely, runtime grows linearly with $\tau$. The performance trend is reasonable: a smaller $\tau$  restricts the information available in the subgraph, whereas an excessively large $\tau$ admits paths from low-confidence rules that degrade the quality of the local embeddings.

\subsection{Analysis of LLM Selection}
\begin{figure}[htbp]
    \centering
    \includegraphics[width=1\linewidth]{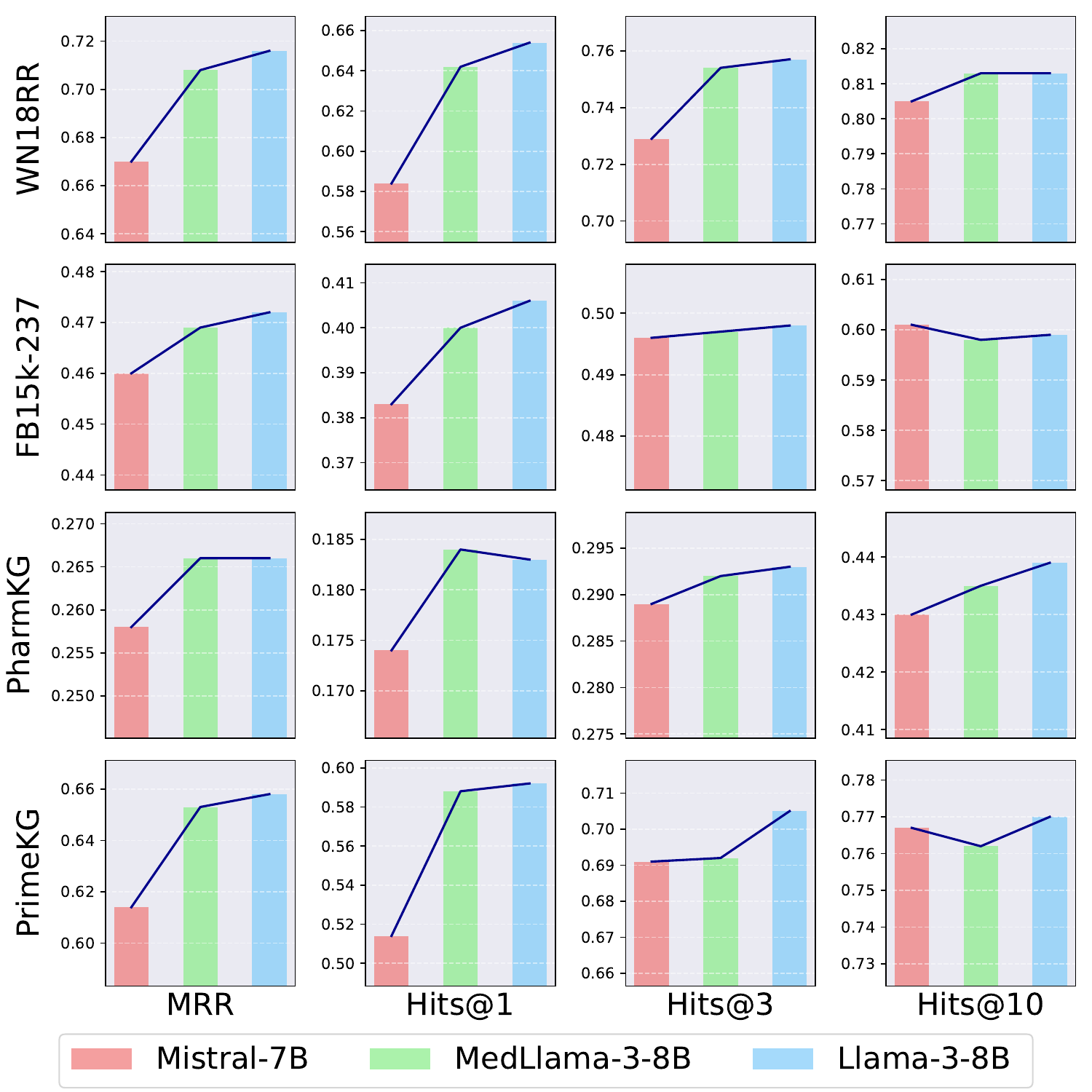} %
    \caption{Comparison of DrKGC performance using different LLMs across four datasets.} 
    \label{fig:model_analysis} 
\end{figure}

In this section, we further investigate the impact of employing other different LLMs within DrKGC on prediction performance. In addition to Llama-3-8B, we compare Mistral-7B and a biomedical-focused instruction-tuning variant, MedLlama-3-8B. The results of replacing the LLM component in DrKGC are presented in Figure~\ref{fig:model_analysis}. Overall, Llama-3-8B delivers the best performance, while Mistral-7B underperforms, despite achieving  the highest Hits@10 on FB15k-237. Notably, MedLlama-3-8B performs slightly worse than Llama-3-8B even on two BKGs, only outperforming in Hits@1 on PharmKG. This suggests that domain-specific LLMs like MedLlama-3, though enriched with biomedical knowledge, may not generalize well to structured relational reasoning tasks such as KGC. MedLlama-3 is primarily optimized for biomedical question answering and clinical text generation, rather than for link prediction or graph-based inference, which limits its effectiveness in this setting. This also demonstrates the benefit of DrKGC’s structure-aware design.

\subsection{Case Study}
To illustrate the practical utility of our approach, we conducted a drug repurposing case study for "breast cancer" using the PrimeKG dataset. In this study, we defined "Breast Cancer" as the query entity and "indication" as the query relation for head prediction. Recognizing that multiple drugs may be effective in treating breast cancer, we employed DrKGC to generate the top 10 predictions. This process was executed iteratively.

To validate our results, we conducted a manual evaluation by clinical trials and published literature \cite{zheng2021pharmkg,xiao2024fuselinker}. Specifically, if a predicted drug is documented on ClinicalTrials.gov, we record the corresponding NCT ID as evidence. If not, we search PubMed for supporting literature and record the corresponding PMID. In the absence of evidence from either source, "No evidence found" is recorded.

\begin{table}[h]
    \centering
    \resizebox{\columnwidth}{!}{%
    \begin{tabular}{llll}  
        \toprule
         & Predicted Drugs & Evidence Source & PMID or NCT ID \\
        \midrule
        1 & Enzalutamide & Clinical Trial & NCT02750358 \\
        2 & Troglitazone & Literature & 31894283 \\
        3 & Rosiglitazone & Clinical Trial & NCT00933309 \\
        4 & Dichloroacetic Acid & Clinical Trial & NCT01029925 \\
        5 & GTI 2040 & Clinical Trial & NCT00068588 \\
        6 & Uridine Monophosphate & Literature & 32382150 \\
        7 & Nimesulide & Clinical Trial & NCT01500577 \\
        8 & Cardarine & Literature & 15126355 \\
        9 & Drospirenone & Clinical Trial & NCT00676065 \\
        10 & Vitamin A & Literature & 34579037 \\
        \bottomrule
    \end{tabular}
    }
    \caption{Top 10 predicted drugs for Breast Cancer.}
    \label{tab:predictBC}
\end{table}

\begin{figure}[htbp]
    \centering
    \includegraphics[width=0.9\linewidth]{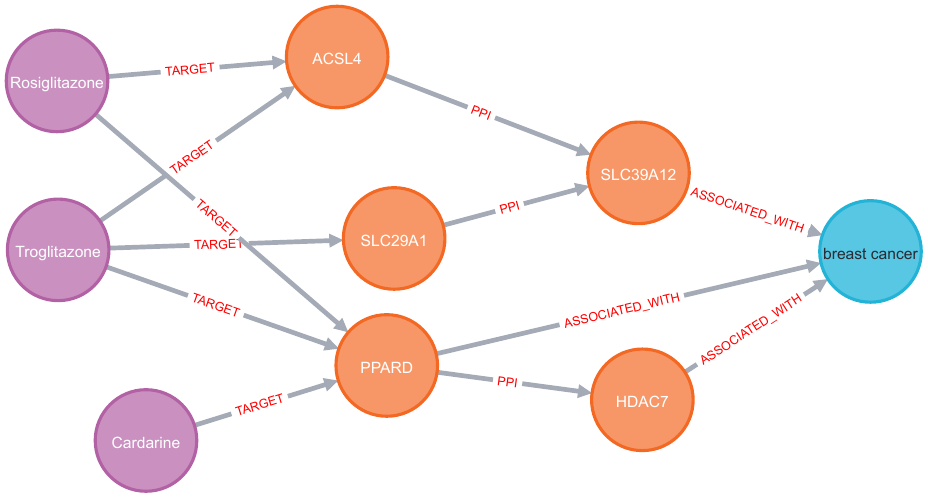} %
    \caption{Example of multi-hop mechanism paths from drugs to Breast Cancer: purple, blue, and orange nodes represent drugs, diseases, and genes/proteins.} 
    \label{fig:case_study} 
\end{figure}


Table~\ref{tab:predictBC} shows the drug repurposing prediction results. Figure~\ref{fig:case_study} illustrates a portion of the retrieved subgraph from the first prediction iteration, where three drugs—Troglitazone, Rosiglitazone, and Cardarine—share mechanism paths that are multi-hop connected to the breast cancer entity. Consider the paths "Troglitazone–PPARD–breast cancer" and "Troglitazone–PPARD–HDAC7–breast cancer": Troglitazone targets PPARD, a druggable protein and a key molecular target in metastatic cancer \citep{zuo2017metastasis}, and PPARD also interacts with HDAC7, which regulates genes critical for tumor growth and the maintenance of cancer stem cells \citep{caslini2019hdac7}. This mechanistic insight provided by DrKGC's subgraph both supports and explains its biomedical predictions.

\section{Conclusion}
In this paper, we propose a novel KGC framework, DrKGC. DrKGC fully exploits graph context information and flexibly integrates mechanisms such as dynamic subgraph information aggregation, embedding injection, and RAG, overcoming the limitations of previous generation-based methods in structural information loss, static entity representations, and generic LLM responses. Experimental results demonstrate that DrKGC achieves state-of-the-art performance on general KGs and performs exceptionally well on domain-specific KGs such as BKGs. By capturing graph context to generate informative subgraphs, DrKGC also enhances model interpretability, which is particularly valuable for biomedical applications.

\section{Limitations}
DrKGC relies on fine-tuning large language models, a process that is computationally intensive, and its performance is inherently constrained by the current capabilities of LLMs and lightweight models. Future work will focus on optimizing fine-tuning efficiency, enhancing LLM performance, and exploring extensions to other graph tasks such as reasoning and question answering. Moreover, retrieving more informative subgraphs may present additional challenges. In this work, we adopt a lightweight heuristic graph retrieval method; however, more rigorous rule-based detection and filtering techniques, as well as alternative subgraph strategies, such as learning-driven subgraph retrieval, merit further investigation. We plan to explore these more sophisticated approaches in future research.

\section*{Acknowledgments}
This work was supported by the National Institutes of Health’s National Center for Complementary and Integrative Health under grant numbers R01AT009457 and U01AT012871, the National Institute on Aging under grant number R01AG078154, the National Cancer Institute under grant number R01CA287413, the National Institute of Diabetes and Digestive and Kidney Diseases under grant number R01DK115629, and the National Institute on Minority Health and Health Disparities under grant number 1R21MD019134-01. The content is solely the responsibility of the authors and does not represent the official views of the National Institutes of Health.

\bibliography{acl_latex}

\newpage
\appendix
\section{Appendix}
\label{sec:appendix}

\subsection{Details of the Dataset}\label{app:dataset}
Table~\ref{tab:datasetstats} presents the statistical details of the four datasets used in our study.

\begin{table}[h]
    \centering
    \resizebox{\columnwidth}{!}{%
    \begin{tabular}{lccccc}
        \toprule
        Datasets   & Entities & Relations & Training  & Validation & Testing \\
        \midrule
        WN18RR     & 40,943   & 11        & 86,835    & 3,034      & 3,134 \\
        FB15K-237  & 14,541   & 237       & 272,115   & 17,535     & 20,466 \\
        PharmKG    & 7,601    & 28        & 400,788   & 49,536     & 50,036 \\
        PrimeKG    & 26,509   & 4         & 130,535   & 500        & 500 \\
        \bottomrule
    \end{tabular}%
    }
    \caption{Statistics of the four datasets.}
    \label{tab:datasetstats}
\end{table}

 WN18RR (\textit{MIT License}), derived from WordNet \citep{miller1995wordnet}, contains word sense entities and lexical-semantic relations like hypernymy. FB15k-237 (\textit{CC BY 4.0}), from Freebase \citep{bollacker2008freebase}, consists of entities such as people and organizations with factual relations like affiliation and location. PharmKG (\textit{Apache-2.0}) focuse on pharmaceutical data, capturing information about genes, diseases, chemicals. PrimeKG (\textit{CC0 1.0}) is a multimodal BKG that unifies other biological entities like phenotypes and pathways for precision medicine analysis. 

For WN18RR and FB15k-237, we adopted the node and relation texts provided by KG-BERT \citep{yao2019kg}. For PharmKG, we utilized the PharmKG-8k version from the original work \citep{zheng2021pharmkg},which filtered high-quality entities based on criteria such as FDA approval and MeSH tree classification and provided a partitioned dataset. 

The PrimeKG dataset used in our study is a subset extracted from the original PrimeKG \citep{chandak2023building} tailored for drug repurposing task. Specifically, we first selected triples from PrimeKG that have a head node of type "drug", a tail node of type "disease", and a relation of "indication". There are 9,388 such triples in total. Next, we randomly split them into 8,388 triples for training, 500 for validation, and 500 for testing, ensuring that the entities in the validation and test sets are also present in the training set. Finally, we enriched the training set by adding additional triples with the following (head, relation, tail) patterns: (drug, target, gene/protein), (gene/protein, associated with, disease), and (gene/protein, ppi, gene/protein). First, we added triples linking the existing drug and disease entities to gene/protein entities; then, we added triples connecting gene/protein entities to one another. In addition, to simplify the problem, we imposed an upper limit on the degree of gene/protein entities to mitigate the influence of hub nodes.

\subsection{Prompt Template}\label{app:prompt_template}

As shown in Table~\ref{tab:prompt_template}, for both the general KG (WN18RR and FB15k-237) and the biomedical KG (PharmKG and PrimeKG), the prompt template remains consistent generally, comprising a simple instruction, a candidates set, corresponding structural embeddings (initially represented by [Placeholder]) for reference, and a question. The only difference is the role name assigned to the LLM (either linguist or biomedical scientist).


\begin{table}[ht]
  \centering
  \begin{adjustbox}{width=\linewidth}
  \begin{tabular}{|p{1.0\linewidth}|}
    \hline
    {\footnotesize\ttfamily\raggedright\sloppy
You are an excellent \{linguist, biomedical scientist\}. The task is to predict the answer based on the given question, and you only need to answer one entity. The answer must be in ('candidate1', 'candidate2', 'candidate3', 'candidate4', 'candidate5', 'candidate6', 'candidate7', 'candidate8', 'candidate9', 'candidate10', 'candidate11', 'candidate12', 'candidate13', 'candidate14', 'candidate15', 'candidate16', 'candidate17', 'candidate18', 'candidate19', 'candidate20').
\\
You can refer to the entity embeddings: 'query entity': [Placeholder], 'candidate1': [Placeholder], 'candidate2': [Placeholder], 'candidate3': [Placeholder], 'candidate4': [Placeholder], 'candidate5': [Placeholder], 'candidate6': [Placeholder], 'candidate7': [Placeholder], 'candidate8': [Placeholder], 'candidate9': [Placeholder], 'candidate10': [Placeholder], 'candidate11': [Placeholder], 'candidate12': [Placeholder], 'candidate13': [Placeholder], 'candidate14': [Placeholder], 'candidate15': [Placeholder], 'candidate16': [Placeholder], 'candidate17': [Placeholder], 'candidate18': [Placeholder], 'candidate19': [Placeholder], 'candidate20': [Placeholder]. 
\\
Question: (The generated question) 
\\
Answer:
    }\\ \hline
  \end{tabular}
  \end{adjustbox}
    \caption{Prompt template for DrKGC}
  \label{tab:prompt_template}
\end{table}




\subsection{Question-Template Lexicon}\label{app:template_lexicon}
For each of the four datasets, two question-template lexicons are provided. One lexicon is designed to use the head node and relation to predict the tail node (corresponding to the tail prediction task), while the other is designed to use the tail node and relation to query the head node (corresponding to the head prediction task). In practice, the appropriate lexicon is selected based on the dataset and the prediction task (head or tail). For each incomplete triple, the corresponding question template is retrieved using the query relation, and then the query entity is inserted into the "\{\}" placeholder, generating the final question. Tables~\ref{tab:tail_prediction_templates_wn} and \ref{tab:head_prediction_templates_wn} illustrate the two question-template lexicons for WN18RR as examples.

\begin{table}[ht]
  \centering
  \begin{adjustbox}{width=\linewidth}
  \begin{tabular}{|p{1.0\linewidth}|}
    \hline
    {\footnotesize\ttfamily\raggedright\sloppy
\textbf{\# tail\_prediction:}\\
\textbf{"also see"}: \\
"What is additionally relevant or similar to \{\}?,"\\
\textbf{"derivationally related form"}: \\
"What is a word or concept that is derivationally related to \{\}?,"\\
\textbf{"has part"}: \\
"What part does \{\} have?,"\\
\textbf{"hypernym"}: \\
"What is a more general category or class that includes \{\}?,"\\
\textbf{"instance hypernym"}: \\
"Of what category or class is \{\} a specific instance?,"\\
\textbf{"member meronym"}: \\
"What is included as a member of \{\}?,"\\
\textbf{"member of domain region"}: \\
"What is associated with \{\} in terms of regional terms or concepts?,"\\
\textbf{"member of domain usage"}: \\
"What is associated with \{\} in terms of specific usage or context?,"\\
\textbf{"similar to"}: \\
"What is similar to \{\}?,"\\
\textbf{"synset domain topic of"}: \\
"What topic or field is \{\} associated with?,"\\
\textbf{"verb group"}: \\
"What verb is in the same semantic or functional group as \{\}?"
    }\\ \hline
  \end{tabular}
  \end{adjustbox}
  \caption{Tail prediction question-template lexicon for WN18RR.}
  \label{tab:tail_prediction_templates_wn}
\end{table}

\begin{table}[ht]
  \centering
  \begin{adjustbox}{width=\linewidth}
  \begin{tabular}{|p{1.0\linewidth}|}
    \hline
    {\footnotesize\ttfamily\raggedright\sloppy
\textbf{\# head\_prediction:}\\
\textbf{"also see"}: \\
"What is related or similar to \{\}?,"\\
\textbf{"derivationally related form"}: \\
"What word or concept leads to \{\}?,"\\
\textbf{"has part"}: \\
"What includes \{\} as a part?,"\\
\textbf{"hypernym"}: \\
"What is a example or specific instance of \{\}?,"\\
\textbf{"instance hypernym"}: \\
"What entity is classified under \{\}?,"\\
\textbf{"member meronym"}: \\
"What larger group does \{\} belong to?,"\\
\textbf{"member of domain region"}: \\
"What is associated with the region of \{\}?,"\\
\textbf{"member of domain usage"}: \\
"What is used in the same context as \{\}?,"\\
\textbf{"similar to"}: \\
"What is considered similar to \{\}?,"\\
\textbf{"synset domain topic of"}: \\
"What is associated with the field or topic of \{\}?,"\\
\textbf{"verb group"}: \\
"What other verb is in the same functional or semantic group as \{\}?"
    }\\ \hline
  \end{tabular}
  \end{adjustbox}
  \caption{Head prediction question-template lexicon for WN18RR.}
  \label{tab:head_prediction_templates_wn}
\end{table}

\subsection{Rule Mining and Subgraph Retrieval Strategy}\label{app:subgraph retrieval strategy}
We first employ the lightweight NCRL model to mine logical rules from the knowledge graph. To further justify the use of NCRL, we evaluate DrKGC by replacing NCRL with RNNLogic and with randomly generated rules. The comparison results are presented in Table~\ref{tab:drkgc_rnnlogic}.

\begin{table}[h]
    \centering
    \resizebox{\columnwidth}{!}{%
    \begin{tabular}{l|cc|cc}
        \toprule
        \multirow{2}{*}{Dataset} 
            & \multicolumn{2}{c|}{\textbf{RNNLogic}} 
            & \multicolumn{2}{c}{\textbf{Random Rules}} \\
        \cmidrule(lr){2-3} \cmidrule(lr){4-5}
            & MRR ($\Delta\%$)     & Hits@1 ($\Delta\%$) 
            & MRR ($\Delta\%$)     & Hits@1 ($\Delta\%$) \\
        \midrule
        WN18RR      & 0.706 (–1.40)        & 0.640 (–2.14)       
                    & 0.682 (–4.75)        & 0.609 (–6.88)        \\
        FB15K-237   & 0.455 (–3.60)        & 0.384 (–5.42)       
                    & 0.446 (–5.51)        & 0.376 (–7.39)        \\
        PharmKG     & 0.265 (–0.36)        & 0.182 (–0.55)       
                    & 0.262 (–1.50)        & 0.180 (–1.63)        \\
        PrimeKG     & 0.652 (–0.91)        & 0.588 (–0.68)       
                    & 0.637 (–3.19)        & 0.569 (–3.89)        \\
        \bottomrule
    \end{tabular}%
    }
    \caption{Performance of DrKGC with RNNLogic and random rules ($\Delta \%$ values indicate differences from using NCRL).}
    \label{tab:drkgc_rnnlogic}
\end{table}

The results show that using rules mined by RNNLogic causes a slight decrease in DrKGC’s performance, demonstrating that the choice of rule mining model can influence overall effectiveness. Employing randomly generated rules leads to a more pronounced degradation and falls behind both the NCRL and RNNLogic, which further validates the appropriateness of NCRL as our rule miner.

To further ensure the quality and reliability of the automatically learned rules, we apply a two-stage post-processing pipeline comprising conflict resolution and redundancy elimination. First, for conflict resolution, we group rules by identical bodies and, when a group yields more than one distinct head, which indicates potential conflict, we retain only the rule with the highest confidence score. Next, to eliminate redundancy, we examine pairs of rules that share the same head: if the body of rule A is a strict subset of the body of rule B and A’s confidence exceeds B’s, we remove B as redundant.

During subgraph retrieval, we constrain the subgraph size by the hyperparameter $\tau$, which limits the number of triples and is set to $100$ after comparing the DrKGC performance of taking $\{50, 100, 200\}$, and control its depth by the length of the rule. The maximum rule length is defined during the training of the logical rule learning model; we set this to 3 to match the configuration of the original NCRL work.

\subsection{Model Training}\label{app:training}
Inspired by previous work \cite{wei2024kicgpt,liu2024finetuninggenerativelargelanguage}, our model training does not strictly follow the traditional paradigm of using fixed training, validation, and test sets. Specifically, we first use the KG dataset’s standard splits for training, validation, and testing to train a lightweight model. We then employ this trained lightweight model to perform head and tail predictions on every triple in the validation set, generating candidate rankings that are used to construct prompts. In the LLM fine-tuning phase, we re-partition the validation set \cite{liu2024finetuninggenerativelargelanguage} and utilize it to fine-tune the LLM. Finally, model performance is evaluated on the test set in the usual manner. For each triple in the test set, both head and tail predictions are conducted to ensure fairness. This approach not only reduces the volume of training data required for fine-tuning but also avoids the issue where the trained lightweight model consistently ranks the correct answer for incomplete triples in the training set first, which could mislead the LLM selection.

In lightweight models training phase, for WN18RR, FB15k-237, and PharmKG, we use the hyperparameters consistent with the original publications of their corresponding methods. For PrimeKG, the optimal parameters identified via grid search are provided in Table~\ref{tab:PrimeKG_hyperparameters}. In the LLM fine-tuning phase, we adjust the learning rate $\{2 \times 10^{-3}, 2 \times 10^{-4}\}$, the number of GCN layers $\{1,2\}$ and the size of GCN hidden dimension $\{128, 256\}$, and set the epoch size to 15 with early stopping. The time required for LLM fine-tuning is detailed in Table~\ref{tab:dataset_time}.

\begin{table}[h]
    \centering
    \resizebox{\columnwidth}{!}{%
    \begin{tabular}{lcccccc}
        \toprule
         & TransE & RotatE & DistMult & ComplEx & R-GCN & HRGAT \\
        \midrule
        Batch Size       & 512 & 512 &  512 & 512  & 256  &  128 \\
        Learning Rate    & 2e-3 & 1e-4 & 1e-4  & 2e-3 & 1e-3 & 1e-3 \\
        Negative Sampling & 512 & 512& 512 & 512 & 512 & 40 \\
        Hidden Dimension & 1000 & 2000 & 2000 & 1000 & 200 & 200 \\
        \bottomrule
    \end{tabular}%
    }
    \caption{Optimal hyperparameters for lightweight model on PrimeKG.}
    \label{tab:PrimeKG_hyperparameters}
\end{table}

\begin{table}[h]
    \centering
    \resizebox{\columnwidth}{!}{%
    \begin{tabular}{lcccc}
        \toprule
         & WN18RR & FB15k-237 & PharmKG & PrimeKG \\
        \midrule
        Runtime & 3:01:31 & 8:21:16 & 2:22:27 & 37:09 \\
        \bottomrule
    \end{tabular}%
    }
    \caption{LLM fine-tuning time statistics.}
    \label{tab:dataset_time}
\end{table}

\subsection{Impact of LLM Size and Generation}\label{app:alternative modules}
In this section, we first evaluate the impact of the LLM's size by comparing Llama-3-8B with a smaller variant, Llama-3.2-3B. The comparison results are presented in Table~\ref{tab:drkgc_llama3_3b}. Overall, the performance achieved with the Llama-3.2-3B LLM is inferior compared to that of the Llama-3-8B, which is consistent with our expectations. This difference arises from the reduced number of parameters in the smaller model, inherently limiting its expressive power and reasoning capabilities.
\begin{table}[h]
    \centering
    \resizebox{\columnwidth}{!}{%
    \begin{tabular}{lcccc}
        \toprule
         Dataset   & MRR ($\Delta$)   & Hits@1 ($\Delta$)  & Hits@3 ($\Delta$)  & Hits@10 ($\Delta$) \\
        \midrule
         WN18RR    & 0.709 (-0.07)   & 0.644 (-0.10)     & 0.754 (-0.03)     & 0.811 (-0.02) \\
         FB15k237  & 0.466 (-0.06)   & 0.397 (-0.09)     & 0.494 (-0.04)     & 0.596 (-0.03) \\
         PharmKG   & 0.260 (-0.06)   & 0.172 (-0.11)     & 0.292 (-0.01)     & 0.436 (-0.00) \\
         PrimeKG   & 0.656 (-0.02)   & 0.595 (+0.03)     & 0.691 (-0.14)     & 0.762 (-0.08) \\
        \bottomrule
    \end{tabular}%
    }
    \caption{DrKGC Performance with Llama-3.2-3B ($\Delta$ values indicate differences from Llama-3-8B).}
    \label{tab:drkgc_llama3_3b}
\end{table}

We also compared with Llama-2-7B, an earlier generation model of similar size. Table~\ref{tab:drkgc_llama2_7b} shows the comparison results relative to Llama-3-8B. 

\begin{table}[h]
    \centering
    \resizebox{\columnwidth}{!}{%
    \begin{tabular}{lcccc}
        \toprule
         Dataset   & MRR ($\Delta$)   & Hits@1 ($\Delta$)  & Hits@3 ($\Delta$)  & Hits@10 ($\Delta$) \\
        \midrule
         WN18RR    & 0.706 (-0.10)   & 0.642 (-0.12)     & 0.749 (-0.08)     & 0.812 (-0.01) \\
         FB15k237  & 0.464 (-0.08)   & 0.396 (-0.10)     & 0.494 (-0.04)     & 0.597 (-0.02) \\
         PharmKG   & 0.262 (-0.04)   & 0.177 (-0.06)     & 0.290 (-0.03)     & 0.436 (-0.00) \\
         PrimeKG   & 0.652 (-0.06)   & 0.587 (-0.05)     & 0.698 (-0.07)     & 0.765 (-0.05) \\
        \bottomrule
    \end{tabular}%
    }
    \caption{DrKGC Performance with Llama-2-7B ($\Delta$ values indicate differences from Llama-3-8B).}
    \label{tab:drkgc_llama2_7b}
\end{table}

\end{document}